# Calibrated Generative AI as Meta–Reviewer: A Systemic Functional Linguistics Discourse Analysis of Reviews of Peer Reviews

Gabriela C. Zapata[1], Bill Cope[2], Mary Kalantzis[2], and Duane Searsmith[2]

*[1]School of Education, University of Nottingham, Nottingham, United Kingdom*

*[2]Department of Education Policy, Organization & Leadership, University of Illinois, Urbana–Champaign and Chicago, United States*

Corresponding author: Gabriela C. Zapata, Gabriela.Zapata@nottingham.ac.uk

# Calibrated Generative AI as Meta–Reviewer: A Systemic Functional Linguistics Discourse Analysis of Reviews of Peer Reviews

This study investigates the use of generative AI to support formative assessment through machine–generated reviews of peer reviews (meta–feedback) in graduate online courses in a public university in the United States. Drawing on Systemic Functional Linguistics and Appraisal Theory, we analyzed 120 meta–reviews to explore how generative AI feedback constructs meaning across ideational, interpersonal, and textual dimensions. The findings suggest that generative AI can approximate key rhetorical and relational features of effective human feedback, offering directive clarity while also maintaining a supportive stance. The reviews analyzed demonstrated a balance of praise and constructive critique, alignment with rubric expectations, and structured staging that foregrounded student agency. By modeling these qualities, AI meta–feedback has the potential to scaffold feedback literacy and enhance learners' engagement with peer review. We conclude that calibrated AI, when embedded in thoughtfully designed workflows, can contribute to hybrid human–AI feedback ecosystems, while also requiring critical attention to its limitations and contextual adaptation in educational practice.

## Introduction

A comprehensive body of literature on formative feedback in higher education (e.g., Evans, 2013; Irons, 2008; Morris et al., 2021) has shown that it is an essential aspect of the learning process, playing both informational and relational roles. This type of feedback, for instance, can not only support students in monitoring progress, refining work, and developing evaluative judgment (Black & Wiliam, 1998; Hattie & Timperley, 2007; Nicol & Macfarlane–Dick, 2006), but can also shape student motivation, trust, and engagement (Hyland & Hyland, 2001; Pearson, 2022). Furthermore, existing studies (e.g., Cho & MacArthur, 2011; Falchikov, 2001; Van Zundert et al., 2010) have shown that tasking students with providing formative comments on their peers' work can also result in enhanced academic performance and improved self–regulatory practices as well as develop their feedback literacy.

Recently, the emergence of generative artificial intelligence (GenAI) has prompted new questions about the ways in which machine–generated comments might expand peer–based formative feedback. In a series of convergent mixed–methods studies, we have explored GenAI–supported formative assessment by piloting successive versions of AI–based feedback systems. In our earlier works, the participating graduate students described peer feedback as more empathetic, specific, and contextually grounded than the AI reviewer's (e.g., Saini et al., 2024; Tzirides et al., 2023, 2024a; Zapata et al., 2024a, 2025a). These findings mirrored those reported in similar studies. For example, Dai et al. (2023) and Steiss et al. (2024) posited that individualized, high–quality GenAI feedback was still difficult to achieve in comparison with that offered by instructors, or, in our case, peers.

Based on the results of these first studies, in our later work, we calibrated the GenAI reviewer with retrieval augmentation, also making use of a more powerful large language model (LLM), GPT–4. These changes positively influenced our participants' perceptions of the AI reviewer, and they increasingly began to value its feedback for its precision, relevance, and actionability (Tzirides et al., 2024c; Zapata et al., 2024b). These findings suggested that fine–tuning AI feedback systems was essential for enhancing their effectiveness.

In our most recent work (Zapata et al., 2025b), we sought to further understand where the differences we had observed in our participants' perceptions of GenAI feedback in earlier studies had originated by focusing on the language used by the GenAI reviewer. To achieve this goal, we employed Systemic Functional Linguistics (SFL) (Eggins, 1994; Halliday & Matthiessen, 2014) and Appraisal Theory (Martin & Rose, 2007; Martin & White, 2005) and analyzed the discourse of non–calibrated and fine–tuned GenAI reviews. The results of the analysis showed how calibration had enabled the system to approximate the rhetorical and relational qualities of effective human feedback by more consistently enacting student agency, balancing critique with encouragement, and modeling academically grounded evaluative moves. These findings suggested that the value of GenAI feedback lay not only in its accuracy and contextual alignment but also in its ability to reproduce the discursive strategies that make formative feedback pedagogically effective.

The present paper extends this line of investigation by focusing on GenAI–generated reviews of peer reviews, an area not yet widely investigated. Specifically, our work seeks to contribute to our knowledge of GenAI–supported feedback in higher education by exploring how this novel form of meta–feedback can support students in developing their capacity as constructive, rubric–aligned reviewers. Drawing on SFL and Appraisal Theory, we analyze how the AI's discourse realizes ideational, interpersonal, and textual meanings, and how these linguistic choices might shape student reviewers' comprehension, confidence, and actionability.

This work is structured as follows. First, we review relevant scholarship. This is followed by the presentation of the theoretical framework that grounded our analysis and the way in which it was carried out. Next, we present and discuss the results of the analysis. In the final sections, we consider pedagogical implications, the study's limitations, and future research directions.

**Literature Review**

Formative feedback is widely recognized as a core component of effective pedagogy, functioning as an instructional intervention that supports learners in closing the gap between current performance and desired goals (Black & Wiliam, 1998; Hattie & Timperley, 2007). Unlike summative assessments, which evaluate learning at a terminal stage, formative feedback is embedded within the learning process, providing opportunities for timely adjustments to instruction and learning strategies. High–quality formative feedback clarifies expectations, evaluates progress toward those expectations, and offers explicit strategies for improvement, aligning with Hattie and Timperley's (2007) three guiding questions: "Where am I going?", "How am I going?", and "Where to next?".

Formative feedback is a central pedagogical practice in higher education, particularly in the development of student writing, as it functions both as a guide for revision and as a means of fostering motivation, self–efficacy, and metacognitive reflection (Holmeier et al., 2018; Nicol & Macfarlane–Dick, 2006; Wiliam, 2011). Research on instructor and peer formative feedback (e.g., Hyland & Hyland, 2001; Morris et al., 2021; Nicol et al., 2014; Pearson, 2022; Van Zundert et al., 2010) has underscored its multidimensional nature, comprising informational, affective, and interpersonal components that work together to assess and support learning.

For instance, Hyland and Hyland's (2001) foundational study of written instructor feedback identified praise, criticism, and actionable advice as the primary discursive functions employed, noting that these were seldom delivered in isolation. Praise, for example, was often used strategically to soften the potentially face–threatening impact of criticism. Hedging devices such as modal verbs ("might," "could"), interrogative syntax ("Is there a better way to express this?"), and personal attribution ("I think…") were also common, contributing to frame negative evaluations in a more palatable form. These strategies thus served dual cognitive and emotional purposes, clarifying the substance of the feedback while preserving rapport with students.

Pearson's (2022) comprehensive review of three decades of instructor formative feedback research reinforced Hyland and Hyland's (2001) findings, also identifying paired–act patterns providing praise and critique within the same sentence as another hallmark of effective practice. Such patterns, alongside the discursive functions noted by Hyland and Hyland, were shown to maintain student motivation and reduce defensive responses. Pearson also developed a typology of common characteristics in written comments, including tone (e.g., advisory, descriptive), syntactic structure, text specificity, and degree of explicitness. Importantly, feedback containing concrete suggestions (e.g., "Consider elaborating on this idea by...") was found to be more actionable and more likely to prompt revision than vague or purely evaluative statements (see also Ferris, 1997; Nurmukhamedov & Kim, 2010). This combination of content critique with clear revision strategies was identified as particularly effective for supporting improvement.

Corpus–based studies have offered more support for these characterizations of instructor feedback. For instance, Lee's (2013) analysis of 126 feedback reports from UK universities revealed modal verbs such as "could," "might," and "would" to be the most frequent markers of hedging, especially in critical or suggestive contexts. These findings align with pragmatic research indicating that mitigation can reduce the interpersonal tension inherent in evaluative discourse (Crismore & Vande Kopple, 1988; Hyland, 1996). In assessment contexts, the strategic use of modal verbs also reflects instructors' ongoing negotiation of their dual roles as both assessors and facilitators—a theme that recurs in feedback scholarship (e.g., see Anson, 1989).

The structural organization of written feedback has also been a subject of close study. For example, Mirador (2000) examined instructor comments on graduate student writing in the UK and identified a recurring six–move rhetorical structure, which she defined as "the clinching pattern," comprising a summary of general impressions, a restatement of the student's ideas, a description of strengths, an identification of weaknesses, suggestions for improvement, and an overall judgment. This structure can offer a coherent narrative that can support both reflection and action. Moreover, as Mirador noted, this organization pattern is not only pedagogically functional but also culturally situated, reflecting genre–specific conventions for communicating evaluative content (e.g., see Swales, 1990; Yelland, 2011).

Another aspect of human feedback that has been explored in the literature is its interpersonal dimension, particularly in terms of engaging students as active participants in the learning process. For instance, Ädel (2018) highlighted the dialogic function of metadiscursive "you" in teacher feedback, showing how it can foster a conversational tone that can clarify expectations and address potential misunderstandings. That is, through direct address and reflexivity, feedback can become an interaction instead of a one–way critique, cultivating a relational ethos that can bolster student agency (Hyland, 2000/2013; Nicol & Macfarlane–Dick, 2006). This interpersonal quality extends beyond teacher–student exchanges to include those between peers. For example, Van Zundert et al. (2010) reviewed research on peer feedback and found that students often respond positively to one another's comments, with reported benefits including enhanced academic performance, increased motivation, and improved self–regulation (see also Falchikov, 2001).

Existing studies on human formative feedback have shown that it constitutes a complex communicative practice that extends beyond the correction of errors. It is characterized by recognizable rhetorical conventions (Mirador, 2000), linguistic strategies of mitigation (Hyland & Hyland, 2001), a dialogic orientation (Ädel, 2018; Van Zundert et al., 2010), and evolving pedagogical best practices (Pearson, 2022). Effective feedback is timely, specific, dialogical, and motivational, aligning with learning objectives while maintaining a supportive tone (Holmeier et al., 2018).

These aspects of formative feedback are particularly crucial when students are tasked with reviewing their peers' work. Existing research (e.g., Bader et al., 2019; Mangelsdorf, 1992), including our own work (Tzirides et al., 2023, 2024a; Zapata et al., 2024a, 2025a), has shown that students' perceived value of peer feedback can vary, often depending on the specificity of the comments (e.g., students tend to reject suggestions that are too general or too short and/or focus only on minor stylistic/grammar issues) and whether they are submitted within the expected time frame or not. With the advent of GenAI, research work has focused on addressing these drawbacks by customizing systems and scaffolds that can offer learners support to offer more effective peer feedback.

For instance, Su et al. (2025) designed and evaluated *Reviewriter*, an LLM–based tool that generates tailored instructions for writing peer reviews, focusing on system design and model fine–tuning. This study involved undergraduate students in computer science courses who were asked to produce peer feedback with and without the system's guidance. Results showed that *Reviewriter* had improved the structure, focus, and constructiveness of peer comments. Similarly, Guo et al. (2025) introduced *EvaluMate*, a chatbot that critiques students' peer–feedback comments in the context of L2 English writing. Working with secondary school students in China, Guo and colleagues collected both peer comments and revisions, finding that the tool of focus had helped students produce more elaborated, specific, and actionable feedback. Students subsequently revised their essays by adding details and examples, suggesting that GenAI can support feedback literacy by shaping reviewing strategies.

Other projects have examined how GenAI can enhance peer feedback at the level of content helpfulness. For instance, Liu et al. (2024) analyzed a dataset of 3,494 rubric–aligned peer review comments from a graduate computer science course to test whether LLMs could automatically evaluate review helpfulness. They proposed a multi–step prompting framework called "Reasoning after Feature Identification," which prompted the model to identify key features of reviews before classifying their helpfulness. This approach outperformed both fine–tuned supervised models and standard prompting methods, showing that GenAI could reliably distinguish helpful from unhelpful reviews without requiring annotated training data.

Collectively, these contributions have shown promising directions for GenAI in improving the usefulness, specificity, and uptake of peer feedback. However, these works have focused only on customization that can optimize usefulness via prompting or tooling and fall short of examining how GenAI feedback is realized through language. That is, none of these studies has explored how the language used by GenAI can enact evaluative stance, open/close dialogic space, scale force/precision, and organize moves that build rapport and prompt action—all of which characterize effective human feedback. Our study addresses this gap by applying SFL and Appraisal Theory to the discourse analysis of AI–generated reviews of peer reviews, unveiling how calibrated GenAI sounds as pedagogy and how its discourse can perform relational and instructional work in higher education. In the following section, we present our study. We begin by outlining the educational context in which our work took place, including the GenAI tool employed. We then present the theoretical framework that guided our analysis and describe the categories applied in it.

## The Present Study

### *Educational Context*

This study was conducted in online graduate courses at a Midwestern U.S. university, with participants enrolled in master's, doctoral, and certificate programs in Education. These courses examined the intersections of learning, technology, and pedagogy, and students developed semester–long projects through iterative cycles of formative assessment. The feedback process was scaffolded by detailed, theoretically grounded rubrics that emphasized theoretical framing, evidence integration, and innovation, ensuring consistency across instructor, peer, and AI contributions (Cope & Kalantzis, 2023b; Tzirides et al., 2023).

The technological infrastructure was provided by *CyberScholar,* a custom–built extension of the *CGScholar* platform (Cope & Kalantzis, 2023a), which was used by the participants to access course materials and develop their work. *CyberScholar* integrates GenAI formative feedback through its *CyberReview* service, which delivers criterion–by–criterion assessments including both a score and qualitative suggestions. To enhance the contextual relevance of feedback, the system uses a Retrieval–Augmented Generation (RAG) pipeline as well as meta–prompts designed to calibrate the AI's evaluative stance and ensure alignment with pedagogical and disciplinary expectations (Tzirides et al., 2024b; Zapata et al., 2024b, 2025b).

In summer 2025, *CyberScholar* introduced a new "feedback on feedback" stage: GenAI reviews of peer reviews carried out by what we called the *AI Assistant*. After students completed their assessments of peers' work, the AI Assistant generated meta–feedback evaluating the quality of those reviews against three rubric–based criteria: *Constructive Advice in the Review*, *Weaknesses of the Review*, and *Strengths of the Review*. For each criterion, the system provided a 0–4 rating and a discursive comment written in a collegial, supportive tone. These meta–reviews were intended to scaffold students' feedback literacy by modeling how peer comments could be made more specific, balanced, and pedagogically effective. The AI's calibration was guided by the results of our previous work, particularly the study examining the discursive characteristics of non–calibrated and calibrated AI reviews of students' work (Zapata et al., 2025b). The initial AI comments and ratings included in the meta–reviews constitute the data for the present study (an example of these reviews, including the meta–prompts given to the AI, is offered in Appendix A). Importantly, in this work, we did not assess the accuracy of the AI's judgments against the underlying student reviews or projects. Our analysis focused on the initial ratings and the language used in the meta–reviews.

### *Theoretical Framework*

This study draws on Systemic Functional Linguistics (SFL), a model of language developed by M.A.K. Halliday (1975, 2009). SFL conceptualizes language as a social semiotic system; i.e., a resource for making meaning that both reflects and shapes social contexts. In contrast to traditional or generative approaches to grammar, which focus primarily on syntactic structures or universal rules, SFL emphasizes how language choices operate in context to simultaneously realize *ideational*, *interpersonal*, and *textual* meanings (Halliday & Hasan, 1985; Halliday & Matthiessen, 2014). These three metafunctions provide complementary perspectives on every utterance, which makes SFL especially well–suited for analyzing discourse in education, where meaning–making is purposeful and situated (Coffin, 2013; Gibbons, 2006; Moore et al., 2018).

The *ideational* metafunction deals with how experience is represented, capturing the way language encodes actions, participants, and circumstances through what is known as the *transitivity system* (Eggins, 1994; Halliday & Matthiessen, 2014). This system allows for analysis of how ideas are built, how agency is assigned, and how phenomena are rendered knowable through linguistic choices. The *interpersonal* metafunction relates to the construction of social relationships and evaluative stance, incorporating grammatical resources that express authority, affect, obligation, and negotiation (Eggins, 1994). The *textual* metafunction, meanwhile, addresses the internal organization of discourse, realized through thematic progression and cohesive devices that contribute to information flow and textual coherence (Eggins, 1994; Halliday & Matthiessen, 2014).

To further investigate interpersonal meaning, we drew on Appraisal Theory, introduced by Martin and White (2005) and later expanded by Martin and Rose (2007). As an extension of the interpersonal metafunction, this framework distinguishes among the evaluation of feelings, people's conduct, and things or texts, the positioning of alternative voices, and the scaling of an evaluation. The distinction among these resources is particularly relevant when the AI evaluates a peer reviewer and the review that student has written. It consists of three subsystems:

- Attitude, which covers *affect* (emotions), *judgment* (moral evaluations), and *appreciation* (aesthetic evaluations);
- *Engagement*, which considers how speakers acknowledge or reject alternative perspectives (monoglossic vs. heteroglossic expressions); and

- *Graduation*, which scales the strength or precision of evaluative language.

Both SFL and Appraisal Theory provide a robust set of tools for analyzing how GenAI–generated feedback communicates meaning to student reviewers. That is, SFL offers a systemic account of clause–level choices and their relationship to broader discourse functions, while Appraisal Theory enables the evaluation of interpersonal tone and persuasive force in feedback.

We argue that this combined framework is particularly appropriate for our research objectives because, as noted earlier, feedback is inherently multifunctional: It conveys information, provides evaluations, organizes discourse, and sustains pedagogical relationships. In the context of GenAI, these functions must be accomplished without shared context or human intuition. Understanding how effectively these functions are performed through linguistic resources is therefore central to evaluating the educational value of AI feedback. For this reason, SFL and Appraisal Theory not only furnish a principled foundation for analysis but also directly address the pedagogical and ethical issues associated with implementing AI in educational contexts (Cope & Kalantzis, 2023c, 2024; Kalantzis & Cope, 2025).

*Data Collection and Analysis*

The dataset for this study comprised the entire corpus of 120 reviews of peer reviews carried out by the AI Assistant during summer 2025 in the participating graduate courses. Each meta–review included three rubric–based components—*Constructive Advice in the Review*, *Weaknesses of the Review*, and *Strengths of the Review*—with a 0–4 rating and a discursive comment per criterion (see Appendix A). This yielded 360 criterion–based evaluation components within the 120 exported records. The corpus excluded later conversational turns between the AI metareviewer and students.

All data were imported into *MAXQDA 2022* (VERBI Software, 2021), which served as the environment for coding, visualization, and pattern analysis. Our coding scheme was derived from the metafunctional categories of SFL (Eggins, 1994; Halliday & Matthiessen, 2014) and the evaluative resources of Appraisal Theory (Martin & White, 2005; Martin & Rose, 2007). Categories were operationalized as follows:

- *Ideational features:* Process types (material, relational, mental, verbal) and circumstantial elements (location, manner/means, cause/purpose). Each clause in the dataset was coded for its process type, with circumstantials tagged to indicate how the AI anchored its suggestions in context.
- *Interpersonal features:* Attitude (judgment, appreciation, affect), engagement (monoglossic vs. heteroglossic formulations), and graduation (scaling of intensity or precision). Segments of text expressing evaluation, stance, or dialogic orientation were coded at the phrase or clause level.
- *Textual features:* Theme–rheme structure, marked themes, cohesion (lexical repetition, conjunction, reference chains), and rhetorical sequencing (e.g., Mirador's [2000] clinching pattern). These categories were applied to larger stretches of text, typically paragraph–level units.

To ensure systematic coverage, all 120 reviews were coded in full. MAXQDA's lexical search and auto–coding functions were used to identify recurring markers (e.g., second–person pronouns, modal verbs, evaluative adjectives), which were then validated manually by the paper's authors. Visualization tools (e.g., code frequency tables) supported the quantitative dimension of the analysis, allowing us to capture proportional distributions of linguistic features in the corpus. Additionally, qualitative memoing was used to annotate excerpts and record interpretive insights about how GenAI enacted feedback functions in context. This integrated procedure enabled us to combine breadth and depth: Quantitative coding provided an overview of linguistic tendencies across all reviews, whereas qualitative annotations traced how these tendencies worked discursively to model rubric–aligned, pedagogically meaningful feedback.

## Results

Our analysis of the 120 GenAI–generated reviews of peer reviews suggests that the system's discourse combined directive clarity with a supportive stance, providing feedback that was not only rubric–aligned but also pedagogically meaningful. That is, the AI enacted a form of meta–feedback that highlighted what reviewers were doing well, where their comments could be improved, and how revisions might be carried out in practice. To capture the multidimensional nature of this discourse, we examined the data through the three SFL metafunctions and Appraisal Theory while also paying attention to how these functions intersected to create feedback that students could both recognize as legitimate and act upon effectively. The results from this analysis are presented as follows, in separate sections.

### *Rubric Ratings in the AI Evaluations*

The 120 AI reviews contained ratings from 0 to 4 for each of the three assessment criterion components. Table 1 shows these scores by criterion. Ratings of 3 predominated in the corpus, but the distribution differed across the three criteria: scores of 2 were more frequent under *Weaknesses* than under *Constructive Advice* or *Strengths*. This suggests that the AI evaluated students' identification and discussion of weaknesses somewhat less favorably than their constructive advice or identification of strengths. This difference also provides context for understanding how the AI balanced positive evaluations with suggestions for improvement in its comments. Importantly, these ratings were assigned by the AI, but do not constitute independent judgments of the quality of the students' peer reviews.

**Table 1**

*Ratings assigned in the AI meta–reviews*

| **Rating** | **Constructive advice** | **Weaknesses** | **Strengths** |
|---|---|---|---|
| 0 | 0 | 1 | 0 |
| 1 | 0 | 0 | 1 |
| 2 | 12 | 25 | 3 |
| 3 | 98 | 94 | 109 |
| 4 | 10 | 0 | 7 |
| **Total** | 120 | 120 | 120 |

*Ideational Metafunction*

At the ideational level, the GenAI reviews represented students as capable, agentive participants in the feedback process, drawing heavily on material processes. As shown in Table 2, over half of all process types in the corpus were material, indicating that the AI Assistant frequently used them to position students as writers engaged in revision, evaluation, and knowledge construction. Comments such as "*You identify specific areas that need improvement such as…"* foregrounded students' actions as reviewers, presenting their contributions as part of a process of identifying problems and proposing improvements. Complementing this action orientation, relational processes evaluated qualities of the reviews themselves, offering characterizations such as "*Your tone is consistently supportive*" or "*The recommendations are feasible*." These evaluations provided students with benchmarks against which their reviews were assessed, indicating not just what they had done but also how their work measured against expected qualities. Mental processes introduced an additional dimension concerned with thinking and understanding. Evaluations such as "*Your review demonstrates a good understanding of…*" also acknowledged students' analytic engagement, although the attribution of understanding in this example was realized through a different grammatical construction. Additionally, while verbal processes were less frequent, they played an important role in situating reviewers' comments within the act of communication, as in "*You point out that more support is needed to…*" or "*You note the absence of supporting data*." The combination of these process types allowed the AI Assistant to evaluate students' reviews while also representing reviewing as an activity involving action, reasoning, evaluation, and communication.

**Table 2**

*Percentage of process types in the corpus*

| Processes | Percentage |
|---|---|
| Material | 51% |
| Relational | 21% |
| Mental | 16% |
| Verbal | 12% |

This ideational orientation was reinforced through circumstantial detail, which anchored the feedback in context and provided student reviewers with clarity about *where*, *how*, and *why* revisions might be made. As Table 3 shows, circumstantials of location were the most common, directing learners to particular sections of their reviews, as in "*For example, in the 'Theory' and 'Concepts' sections….*" Additionally, circumstantials of manner/means clarified how improvements might be achieved, as illustrated by "*This is a good start, but you could go further by identifying how those theories might reframe the argument.*" Here, the expression "*by identifying*" specifies the means through which the suggested improvement could be accomplished. Circumstantials of cause/purpose, in turn, explained the intended outcomes of particular suggestions, as in "*To reach the highest level of this criterion, aim to consistently pair each critical observation with a specific and actionable revision strategy….*" These circumstantial resources contributed to the specificity of the feedback by connecting evaluations with textual locations, possible revisions, and their purposes. The ideational findings show how the AI Assistant represented students as active participants in the reviewing process while situating their work within evaluative and procedural frames that oriented them toward revision.

**Table 3**

*Percentage of circumstantials in the corpus*

| **Circumstantial elements** | **Percentage** |
|---|---|
| Location | 40% |
| Manner/Means | 32% |
| Cause/Purpose | 28% |

*Interpersonal Metafunction*

The interpersonal analysis showed that the AI Assistant maintained a balance between support and critique, constructing an evaluative stance that was both collegial and academically oriented. As summarized in Table 4, positive judgment and appreciation were nearly universal, appearing in comments such as "*Your review demonstrates a strong effort to guide the writer with thoughtful and aligned suggestions*." These affirmations acknowledged the value of students' work while reinforcing the qualities expected of effective peer reviews. Their placement alongside critical comments also contributed to an evaluative structure in which the identification of strengths provided a context for introducing suggestions for improvement.

Importantly, critique itself was not absent, but it was frequently expressed through negative judgment framed in ways that moderated its force. For example, statements such as "*Your comments align well with the rubric criteria… that said, there are a few opportunities to enhance the constructiveness of your advice further*" illustrate how the AI Assistant moved from positive evaluation to critique without dismissing students' existing contributions. The expression "*a few opportunities*" limited the scope of the criticism, while "*further*" presented improvement as an extension of work already recognized as valuable. Therefore, the feedback combined acknowledgment of students' achievements with the identification of areas requiring further development.

**Table 4**

*Percentage of appraisal categories across all reviews*

| Category | Percentage of reviews |
|---|---|
| Positive Judgment & Appreciation | 92% |
| Judgment– (calibrated critique) | 79% |
| Appreciation– | 58% |
| Affect+ (positive affect) | 75% |
| Affect– (negative affect) | Rare |

Alongside judgment, the feedback frequently drew on appreciation, with the AI

Assistant praising qualities of the reviews themselves, such as clarity, tone, or specificity. While judgment evaluated students' demonstrated capacities and reviewing practices, appreciation focused on the qualities and value of the reviews as products. Positive affect also appeared in the corpus, contributing to the overall supportive stance. Expressions such as "*Thank you for your thoughtful and diligent review*," often found at the beginning or end of reviews, further personalized the interaction through conventional forms of interpersonal politeness. The rare presence of negative affect also contributed to a pattern in which criticism was primarily conveyed through evaluations of students' work rather than expressions of negative emotion.

Beyond attitude, the reviews also featured engagement resources that introduced alternative perspectives and qualified evaluations, creating opportunities for further interpretation. These resources were evident when the AI Assistant acknowledged a student's existing observations before suggesting how the critique could be developed further, as illustrated in the following example:

> "Similarly, in your comments on theory and reasoning, you do begin to highlight underexplored areas—such as the lack of connection between Deci's SDT and the Chinese study—but your critique would be stronger if you more explicitly pointed out missed opportunities for synthesis or more rigorously examined logical gaps."

In this example, the AI Assistant first acknowledged the student's contribution through "*you do begin to highlight*," before introducing a qualification with "*but*." The conditional construction "*would be stronger if*" then presented the suggested improvement as a possibility rather than an absolute judgment. The evaluation thus recognized what the student had already accomplished while identifying how the critique could be expanded. This combination of acknowledgment and qualification shows how the AI Assistant used engagement resources to manage evaluative stance and create space for alternative ways of developing the review.

Also important was the use of scalar modifiers and modal expressions to highlight success and soften critique while providing precise guidance. For example, expressions such as "*Your critique would be even more helpful if…*" conveyed positive evaluation through "*even more*," while simultaneously introducing an additional expectation. Formulations such as "*This could be taken further by…*" used modality to present revision as a possibility instead of a mandatory step for improvement. Although these resources perform different grammatical and evaluative functions, their combination allowed the AI Assistant to express degrees of achievement and formulate suggestions without relying exclusively on direct commands. Therefore, the resulting discourse maintained an academic but approachable tone, presenting feedback as guidance for further development.

A final interpersonal aspect worth noting was the AI Assistant's use of the second–person pronoun, which positioned the student as an agentive participant in the reviewing process and created a direct, dialogic tone, even in a written review. Also, the employment of "*you*" contributed to the supportive tenor by addressing students directly and foregrounding their role in evaluating and revising their comments. For example, saying "*You might consider…*" is less authoritative than using an imperative such as "*Add…*" or an impersonal construction such as "*It should be added….*" In other words, the combination of second–person address and modal expressions presented students as individuals capable of making decisions about their reviews, while also allowing the AI Assistant to communicate expectations for improvement. The interpersonal results show how the GenAI reviews enacted rhetorical and relational qualities associated with effective human feedback, combining encouragement with critique and presenting revision as an opportunity for further development.

*Textual Metafunction*

Textually, the GenAI reviews were coherent, rhetorically patterned, and structured in ways that contributed to the clarity of the feedback. Marked Themes consistently introduced new evaluative moves, orienting students to the focus of each comment. For example, the marked topical Theme in "*With slightly more detailed alignment to rubric language…*" established the conditions under which the student's review could be improved, preparing the reader for the evaluation that followed. Cohesion was maintained through features such as lexical relationships (e.g., the use of "*review*," "*evaluation*," and "*critique*" to refer to related aspects of students' reviews), referential ties (e.g., "*You might consider briefly acknowledging… **Doing so** would…*"), a variety of conjunctions/connectors (e.g., "*However*," "*Overall*," "*Finally*"), and the employment of demonstrative pronouns and determiners (e.g., "***These** are helpful and align well*" and "*Thank you for your thoughtful and diligent review…I appreciate the collegial tone and the clear effort you have put into **this** analysis*"). These resources established connections among successive evaluations and suggestions, allowing the AI Assistant to maintain a consistent focus on the students' reviews while moving between different aspects of their performance. Also, the feedback frequently followed the *Given → New information* principle, acknowledging what students had done before introducing areas for improvement, as in "*You highlight strengths in the author's communication style… This could be expanded by specifying which theoretical connections were most effective*." Here, the initial acknowledgment of the student's observations provided the point of departure for introducing a more specific suggestion. This progression contributed to the coherence of the feedback by connecting new recommendations to aspects of the reviews that had already been evaluated.

In addition to thematic progression and cohesion, the AI Assistant made systematic use of interpersonal resources that both managed its interaction with student reviewers and contributed to textual organization. Although these resources primarily enacted interpersonal meanings, they also helped organize the unfolding discourse by marking openings, transitions, and conclusions. One important set of features involved speech roles and functions, which appeared in every review analyzed.

Commands in structures such as "*To more effectively highlight strengths, consider being more explicit in linking those strengths to the stated goals of the creator's project or to the broader implications of their work*" framed feedback as directive yet constructive. The imperative "*consider*" presented the suggested action as something the student could examine rather than as an obligation, while the initial purpose expression established the intended outcome of the recommendation. Alongside these, offers like "*Let me know if you'd like a deeper explanation or examples—I'm happy to help further!*" extended support and presented the AI Assistant as a cooperative partner. Particularly notable was the employment of offers realized through declarative and imperative structures containing conditional clauses. For instance, expressions such as "*Happy to elaborate further if you'd like to continue the conversation!*" or "*Please feel free to reach out if you'd like a more detailed breakdown or examples to refine your review further*" conveyed willingness to provide additional assistance while also functioning as textual markers. These expressions frequently appeared toward the end of feedback segments, signaling closure while maintaining a collegial stance. Therefore, their contribution can be said to have been both interpersonal and textual: they represented further interaction as possible while helping establish recognizable boundaries within the reviews.

The reviews also frequently employed vocative adjuncts, which appeared in 98% of cases. Openings such as "*Hi [student's name]*," or "*Dear [student's name]*," and closings like "*Great work, [student's name]!*" personalized the interaction and contributed to a collegial relationship between the AI Assistant and student reviewers. These vocatives also performed a textual function by marking the beginning or conclusion of evaluative sequences. Their placement helped establish boundaries within the discourse, while their interpersonal meaning maintained the direct address characteristic of the reviews. Finally, the use of the second–person pronoun "*you*" as topical Theme was nearly ubiquitous, appearing in 98% of reviews. Clauses such as "*You do offer advice in areas where the work can be improved*," "*You do a* *good job*," and "*You demonstrate a clear understanding of...*" consistently placed the student reviewer in clause–initial position, thematically orienting the discourse around their actions and contributions. This repeated use of "*you*" sustained a cohesive referential focus by maintaining the student reviewer as the point of departure for successive evaluations and suggestions. Additionally, it contributed to the progression of information by connecting what students had accomplished with what they might subsequently improve. In this way, the textual structure of the reviews was closely connected to their interpersonal stance, keeping students at the center of the feedback while organizing the evaluations and recommendations around their reviewing practices.

In terms of rhetorical organization, the reviews exhibited features associated with Mirador's (2000) clinching pattern, moving from summary, to strengths, to weaknesses, and concluding with overall judgments. For example, reviews began with affirmation ("*Thank you for your thoughtful and diligent review*"), proceeded to critique ("*Your review is respectful, but the depth of analysis could be enhanced*"), and ended with a final evaluation and rating. This organization established a recognizable sequence in which positive evaluations preceded the identification of limitations and suggestions for improvement. The final judgments and ratings brought these evaluative moves together, providing closure to each criterion–based assessment. Such patterned organization made the progression of the feedback explicit, allowing students to distinguish among the acknowledgment of strengths, the identification of weaknesses, and the recommendations for further development.

In sum, the findings suggest that the GenAI reviews of peer reviews were linguistically sophisticated and organized in ways consistent with the discourse of formative feedback. Ideationally, they foregrounded student agency while providing detailed circumstantial anchoring; interpersonally, they combined praise with calibrated critique; and textually, they employed cohesive and rhetorical strategies that made the feedback coherent and its recommendations explicit. The analysis also showed how linguistic resources associated with one metafunction could contribute to meanings associated with another. For example, vocatives, offers, and second–person topical Themes simultaneously shaped the interpersonal relationship between the AI Assistant and student reviewers and contributed to the organization of the discourse. The results illustrate how GenAI, guided by meta–prompts and rubrics, can reproduce rhetorical and relational features associated with effective human feedback. Although the study did not examine students' responses to the meta–reviews, these linguistic characteristics provide a basis for considering how AI–generated feedback might contribute to the development of feedback literacy.

## Discussion

Our findings suggest that a calibrated GenAI reviewer can reproduce key discursive qualities associated with effective human feedback and, crucially, apply them at the level of meta–feedback (i.e., reviews of peer reviews), potentially contributing to students' feedback literacy within a rubric–governed workflow. In the 120 meta–reviews we analyzed, the AI consistently represented student reviewers as capable agents (via material processes), grounded its suggestions in concrete circumstances (location, manner/means, cause/purpose), and paired these ideational choices with an interpersonal stance that balanced praise with calibrated critique while maintaining a dialogically oriented tone through hedging, modality, and offers of further help.

Textually, the feedback was cohesive and rhetorically patterned (e.g., summary → strengths → weaknesses → suggestions → closing judgment), and it employed interpersonal resources such as speech functions, vocatives, and second–person topical Themes, not only to manage stance and participation but also to mark boundaries, transitions, and closure, all of which contributed to textual coherence. More importantly, through the adoption of these discursive features, the meta–reviews did more than assess peer comments: they modeled ways of reviewing by making expectations visible, pointing to where and how a review could be strengthened, and maintaining a collegial tone that invited further engagement. The combination of these features allowed the AI Assistant to communicate both an evaluation of students' reviews and suggestions for developing their feedback practices.

These findings extend and help explain shifts we reported in earlier phases of our program of research. In our first studies (e.g., Saini et al., 2024; Tzirides et al., 2023; Zapata et al., 2024a, 2025a), we found that students tended to value peer feedback over generic AI commentary, citing empathy, contextual awareness, and dialogic nuance as advantages of human reviews. However, after introducing retrieval augmentation and calibration, students increasingly rated GenAI feedback as relevant, precise, and actionable (Tzirides et al., 2024b; Zapata et al., 2024b). By focusing here on GenAI reviews of peer reviews, our analysis clarifies how those perceived gains were realized linguistically: The AI's preference for action–oriented transitivity, its systematic use of circumstantials to anchor suggestions, and its calibrated interpersonal stance together yielded feedback that felt specific and usable while remaining supportive. In other words, the improvements documented after RAG and calibration can be traced to recurring linguistic choices that operationalized rubric expectations in discourse (Zapata et al., 2025b) and might have rendered them teachable for student reviewers. And while our previous work examined how calibration influenced students' perceptions and the discourse of AI–generated reviews of their work (Zapata et al., 2025b), the present study extends that analysis to feedback directed at students in their role as reviewers. We believe this is an important and novel distinction since the object of evaluation is now the students' own feedback practices instead of the academic work they produced.

The interpersonal profile of the meta–reviews also aligns closely with established scholarship on formative feedback. For example, Hyland and Hyland (2001) described effective comments as combining praise, criticism, and actionable advice, often using mitigation to preserve rapport, while Pearson (2022) emphasized "paired acts" that nest critique within affirmation to sustain motivation. We see these dynamics linguistically instantiated here through the AI Assistant's positive judgment and appreciation alongside calibrated critique, scaled by graduation resources. Moreover, the AI's dialogically oriented discourse, realized through hedges, modal expressions, and explicit offers to elaborate, resembles features of the feedback conversations described in work on metadiscourse and engagement (e.g., Ädel, 2018). Notably, these interpersonal moves also performed textual functions. Vocatives helped frame openings and closings, whereas offers and softened directives contributed to the organization of evaluative sequences, allowing student writers to recognize transitions between assessment, suggestions, and concluding remarks. The interpersonal stance of the AI Assistant was, therefore, closely connected to the textual organization of the feedback, with linguistic resources contributing simultaneously to the construction of a collegial relationship and the coherence of the reviews.

Additionally, the textual organization we observed in the AI's use of marked Themes to introduce evaluative moves, cohesive ties across propositions, and consistent *Given → New* staging echoes features associated with high–quality academic writing and scholarly discourse (Eggins, 1994; Swales, 1990; Yelland, 2011), as well as rhetorical structures reported in research on written instructor feedback, including Mirador's (2000) "clinching pattern." In our corpus, this sequencing constituted a recurrent organizational pattern through which the AI presented rubric–based evaluations in a way that distinguished what students had done well, where improvement was needed, why particular changes were suggested, and how they might proceed. Also, by sustaining second–person "*you*" as topical Theme and maintaining vocative anchoring, the system kept the peer reviewer as a continuous point of reference, which both personalized the feedback and contributed to referential cohesion. As in our previous work on calibrated AI feedback to student drafts (Tzirides et al., 2024b; Zapata et al., 2024b, 2025b), the linguistic characteristics of these reviews reflected the evaluative expectations incorporated into the system's design. That is, the meta–prompts explicitly requested supportive, balanced, and constructive feedback, and our analysis shows how these expectations were realized through particular choices in transitivity, Appraisal, Theme, and rhetorical organization. Thus, the contribution of the present analysis is to explain how these specifications were expressed linguistically, and not to establish the independent effect of calibration, which would require comparing different system configurations.

Positioned within the broader literature on AI–supported peer assessment, these results contribute two clarifications. First, studies that use prompting or tool scaffolds to improve the helpfulness of peer comments (e.g., Guo et al., 2025; Liu et al., 2024; Su et al., 2025) have shown that GenAI can support the structure and specificity of feedback. Our findings complement these works by examining how a calibrated system realizes evaluative and interpersonal meanings linguistically when the object of assessment is the peer review itself. Second, our analysis suggests that meta–feedback can be understood as a potential resource for developing feedback literacy. That is, by modeling rubric–aligned reviewing moves in discourse, the AI can make instructor expectations more explicit and provide students with examples of how peer comments might be deployed. The “feedback on feedback” stage in *CyberScholar* extends earlier human–versus–AI debates by considering GenAI as a pedagogical partner whose role is to make disciplinary criteria visible and provide guidance on reviewing practices, instead of replacing human judgment. Whether students appropriate these models in subsequent feedback tasks remains a question for further investigation.

There are, however, important boundaries to note. As we have argued in our prior studies (e.g., Zapata et al., 2025a, 2025b), AI systems do not share the histories and situated interpersonal experiences that peers and instructors bring to feedback, even when calibrated. While the meta–reviews in this work enacted a supportive tone and offered dialogic openings, these linguistic characteristics cannot be equated with the relational depth of human interaction. A potentially valuable role for GenAI in this context is therefore complementary: to provide consistent and rhetorically coherent models that students can examine, adapt, and question in conversation with peers and instructors. Designing for that role would involve continuing to consider calibration through meta–prompts tied to rubrics and course discourse, retrieval grounded in relevant materials, and interfaces that allow students to critically engage with the feedback they receive.

Finally, the present analysis also makes a methodological contribution by showing how SFL and Appraisal Theory can be employed to examine the specific linguistic resources through which AI–generated feedback constructs evaluative and interpersonal meanings. This perspective provides a connection between system design and discourse analysis, as it allows us to examine how expectations incorporated into meta–prompts and rubrics are realized in the language of the resulting feedback. For example, material processes and circumstantial elements can be examined in relation to action–oriented suggestions, whereas Appraisal resources and thematic organization can provide information about how evaluations are expressed and organized. These analytical categories offer researchers a theoretically grounded means of describing AI feedback beyond general assessments of its helpfulness or coherence. And they can also inform further examinations of the relationship between the design of GenAI feedback systems and the discourse those systems produce.

**Pedagogical Implications**

This study's findings point to several implications for the design of formative assessment in higher education. First, the evidence that GenAI reviews of peer reviews can consistently foreground student agency, provide circumstantially anchored suggestions, and employ a supportive interpersonal stance suggests that such systems may play a role in scaffolding feedback literacy. That is, by modeling how reviewers can combine praise with constructive critique, sequence strengths before weaknesses, and ground observations in rubric criteria, the AI can make evaluative expectations explicit in its discourse. This might be particularly relevant given that students often report uncertainty about what constitutes "good feedback" (Nicol et al., 2014; Van Zundert et al., 2010). The AI's meta–feedback offers concrete examples of reviewing practices that learners could examine and potentially draw upon in future feedback tasks. Although our analysis did not investigate whether students subsequently incorporated these practices into their own reviews, the linguistic characteristics identified illustrate how such modeling is realized in AI–generated discourse.

Second, *CyberScholar*'s workflow, which combines instructor, peer, and AI input, illustrates the potential of hybrid feedback ecosystems. Our results show how AI–generated meta–feedback can function as an additional evaluative layer within these systems by expressing rubric expectations, modeling reviewing practices, and providing suggestions for improvement. These functions could complement the interpretive, empathetic, and context–sensitive contributions of peers and instructors. GenAI could therefore operate as a form of "cyber feedback" (Cope & Kalantzis, 2023c), expanding human assessment practices without replacing them (Zapata et al., 2024a, 2025a). The contribution of the present study is to describe the possible discursive characteristics of this additional feedback layer and how they could relate to the formative purposes of the assessment workflow.

Finally, the dual interpersonal and textual role of features such as vocatives, hedging, and offers illustrates how linguistic calibration is relevant to the design of educational feedback. The system's tone and rhetorical patterning contributed to the construction of a supportive evaluative stance while also organizing the discourse in ways that enhanced coherence. This implies that designing AI feedback systems involves considering not only the accuracy of evaluations or their alignment with rubrics, but also the linguistic resources through which feedback is communicated. Pedagogically, instructors could use AI–generated meta–reviews as teaching resources by, for instance, asking students to compare GenAI reviews with their own, analyze how discursive moves align with rubric criteria, or critically evaluate the AI's balance of praise and critique. These activities would provide opportunities for students to examine the relationship between linguistic choices and the purposes of formative feedback. However, their effectiveness in supporting feedback literacy still needs to be investigated through research focusing on students' engagement with and use of these resources.

**Limitations**

Despite these promising implications, the study has limitations that warrant consideration. First, while the corpus of 120 GenAI reviews provides evidence of recurring linguistic patterns, our analysis was limited to one institutional context and a set of courses within a specific academic program. The extent to which these findings generalize to other disciplines, levels of study, or institutional settings remains uncertain, particularly given that disciplinary discourse norms vary widely (Hyland, 2000/2013). Our analysis describes the discourse produced within a particular AI–supported assessment workflow and should be understood in relation to that context.

Second, while our use of SFL and Appraisal Theory enabled fine–grained linguistic analysis, this approach does not directly capture students' perceptions of or responses to the AI feedback. Although prior studies in our program of research have documented student perspectives (e.g., Saini et al., 2024; Tzirides et al., 2023, 2024a, 2024b; Zapata et al., 2024a, 2024b, 2025a), this specific study analyzed AI discourse itself. We did not examine whether students accepted, questioned, or incorporated the suggestions into subsequent reviewing practices. Further research could connect linguistic findings with learner reflections to better understand how such meta–feedback is interpreted and used. The absence of these data does not affect our analysis of the linguistic characteristics of the meta–reviews, but it limits the conclusions that can be drawn about their effects on students' feedback literacy.

Third, while calibration through meta–prompts and RAG can contribute to the specificity and contextual alignment of AI discourse, these strategies remain technically and contextually bounded. The AI does not have the shared histories, classroom experiences, or interpersonal knowledge that shape peer and instructor feedback. Although the system produced linguistic features associated with supportive and collegial interaction, this does not establish that it reproduced the affective and relational dimensions of human dialogue. Additionally, because our study examined the discourse generated by one configured AI system, we cannot isolate the effects of its calibration procedures from other aspects of its design. Our findings describe how the system realized evaluative and interpersonal meanings within the context investigated, while our previous research provides the broader basis for considering the role of calibration in AI–supported formative assessment.

**Conclusion**

This study contributes to the growing literature on AI in education by examining how GenAI can support formative assessment through the practice of reviews of peer reviews. Our analysis of 120 AI-generated meta-reviews showed that the system consistently realized ideational, interpersonal, and textual meanings in ways associated with effective feedback practices, positioning students as active agents, offering contextually grounded suggestions, balancing praise with critique, and structuring discourse coherently. These findings extend prior research on AI feedback by examining a different object of evaluation and demonstrating how linguistic resources contribute to the construction of rubric-aligned, supportive, and action-oriented feedback. The analysis also shows the value of examining GenAI feedback as discourse, complementing studies that focus primarily on the accuracy, relevance, or perceived helpfulness of machine-generated evaluations.

The implications are twofold. Pedagogically, GenAI meta-feedback may provide a resource for scaffolding feedback literacy by offering students examples of how to review peers' work in rubric-aligned, constructive, and dialogically oriented ways. Although the present study did not examine students' use of these examples, the discourse analyzed illustrated how the AI Assistant made evaluative expectations visible through its linguistic choices. Conceptually, our analysis demonstrates the value of treating AI feedback as discourse, where linguistic resources contribute to the construction of evaluative meanings, interpersonal relationships, and textual organization. Examining these resources allows us to describe what the AI communicates, how it communicates it, and how its discourse might relate to the formative purposes of the assessment context.

Future research could build on this foundation by investigating how students respond to and integrate GenAI meta-feedback into their own reviewing practices, by examining similar systems in diverse disciplinary contexts, and by exploring how different calibration strategies shape the discourse of AI-generated feedback. Such investigations would complement the linguistic analysis presented here and contribute to a better understanding of the role GenAI might play in formative assessment. Within this broader educational context, AI-generated meta-feedback can be considered a potential complement to human evaluation, offering students additional opportunities to examine and develop the discursive practices involved in constructive peer reviewing.

**Appendix A**

*Sample GenAI Review of a Student's Peer Review*

**I. Criterion: Constructive Advice in the Review**

Please assess this aspect of the review by giving it a rating of 0 to 4. As you make your assessment, please consider these important characteristics of a helpful reviewer. Please address the reviewer by their name and refer to yourself as CyberReviewer.

Their review should be:

- Constructive, offering actionable feedback to the creator of the work.
- Consistent, ensuring that suggestions for revision align with the reviewer's critical comments.

Please be clear about the specific ways the reviewer can strengthen their review. Focus mainly on the review but also look at the work they are reviewing in case there are important things the reviewer may have missed. Suggest ways the review could be more helpful to the creator of the work.

***AI Assistant's review:***

Dear [student's name],

Thank you for your thoughtful and well–structured review. As CyberReviewer, I appreciate the clarity and scholarly tone you bring to your evaluation. Your review consistently highlights the strengths of the author's work and provides commendations across all criteria. However, in assessing the criterion of Constructive Advice in the Review, I found that while your feedback is supportive and affirming, it could be more helpful to the creator if supplemented with actionable suggestions.

You accurately praise the author's strong use of educational theory, empirical evidence, and conceptual clarity, and you provide concise summaries of what the author does well. However, the review would be significantly more constructive if you had also included specific, aligned recommendations for further improvement or elaboration. For example, while you note that ethical concerns are "briefly but thoughtfully mentioned," it would be helpful to suggest expanding on these issues with concrete examples or frameworks, such as algorithmic fairness or culturally responsive AI design.

Similarly, in your comment on the communication criterion, you praise the "lucid and accessible" writing and use of subheadings, but offering a suggestion—such as how to further integrate or contextualize embedded media—would make your feedback more actionable. Your comments align well with the rubric, but the lack of targeted suggestions for revision, even in areas that are already strong, limits the developmental value for the knowledge creator. A more balanced review would not only celebrate existing strengths but also gently push the creator toward refinement and depth.

**Rating: 3**

You've shown significant effort and a strong grasp of the content and criteria, [student's name]. To reach the highest level of effectiveness, I encourage you to include at least one specific, constructive suggestion per criterion. I'd be happy to elaborate further if you'd like to workshop one of your comments together.

**II. Criterion: Weaknesses of the Review**

Please highlight the weaknesses of this review. Please assess this aspect of the review by giving it a rating of 0 to 4. An excellent review will be gently critical. It will also be:

- Balanced, highlighting weaknesses of the work under review while at the same time comparing these with its strengths.
- Respectful, even if the reviewer disagrees with the author's argument or conclusions.
- Focused, confined to the review criteria provided by the rubric agents.

Suggest ways the review could more persuasively highlight weaknesses to the creator of the work. Focus mainly on the review but also look at the work they are reviewing in case there are important things the reviewer may have missed. Please address the reviewer by their name and refer to yourself as CyberReviewer.

Check for content alignment between each review comment and the intended rubric criterion they are addressing. If a misalignment is noticed, advise the reviewer to revise the content or its placement in the review to improve clarity and coherence.

Evaluate if the total word count of the review meets the expectations set by the course instructor and ensure reviewers are reminded to follow the specified word count limits outlined in the course instructions.

***AI Assistant's review:***

Dear [student's name],

Thank you for your thoughtful and well–structured review. As CyberReviewer, I appreciate the clarity and professionalism you bring to your feedback. In evaluating your review under the criterion "Weaknesses of the Review," I find that overall, your review demonstrates a commendable level of effort and attention. However, there are areas where it could be strengthened to better meet the expectations of a gently critical and analytically balanced review.

You consistently award the reviewed work a perfect score across all criteria, but your comments rarely address any weaknesses or limitations in the creator's work. For example, while you highlight the strengths in theoretical explanation and referencing, you do not raise any questions about potential overreliance on theory, under–exploration of counterarguments, or the need for more critical discussion of implementation challenges. Similarly, under "Evidence," you acknowledge the use of statistics and peer–reviewed sources but do not comment on whether there are gaps in data, such as lack of longitudinal studies or overreliance on U.S.–centric data.

A key opportunity for improvement would be to incorporate more balanced critiques. For instance, under “Application,” you might ask: Are the proposed AI applications feasible in under–resourced schools? Or under “Innovation,” you might question whether the ideas proposed are fully original or mostly build upon existing models. These questions would gently challenge the creator to refine their work while maintaining a respectful tone.

Lastly, while your comments are concise, they often lack the depth needed to fully justify a top score of 4. Expanding each comment by 1–2 sentences to include constructive critique would enhance both clarity and usefulness.

**Rating: 2**

Let me know if you’d like a breakdown of how to rephrase or expand any specific comments. I’m happy to help you enhance your review further.

Best regards,

CyberReviewer

**III. Criterion: Strengths of the Review**

Please assess this aspect of their review by giving it a rating of 0 to 4. As you make your assessment, please bear in mind that an excellent review will be:

- Supportive, written in a collegial and seriously minded tone.
- Congratulatory, highlighting the strengths of the review in terms of the way it addresses the review criteria provided by the rubric agents.

Suggest ways the review could more effectively highlight strengths of the work it is reviewing. Focus mainly on the review, but also look at the work they are reviewing in case there are important things the reviewer may have missed. Please address the reviewer by their name and refer to yourself as CyberReviewer.

Check for content alignment between each review comment and the intended rubric criterion they are addressing. If a misalignment is noticed, advise the reviewer to revise the content or its placement in the review to improve clarity and coherence.

Evaluate if the total word count of the review meets the expectations set by the course instructor and ensure reviewers are reminded to follow the specified word count limits outlined in the course instructions.

***AI Assistant's review:***

Dear [student's name],

Thank you for your thoughtful and detailed review. As CyberReviewer, I'd like to commend you on the supportive and collegial tone you maintained throughout your feedback. Your review reflects a serious and respectful engagement with the author's work, and your comments are clearly grounded in the rubric criteria. You consistently highlighted strengths across all dimensions of the rubric, from conceptual clarity to practical application and innovation, demonstrating a strong grasp of both the assignment and the work under review.

You did an excellent job identifying specific strengths in the creator's writing and argumentation. For example, your comment on the "lucid and accessible" academic writing, and your recognition of the "hopeful but measured tone" in the application section, are clear and affirming. These kinds of observations give the creator confidence in what they are doing well and encourage them to continue building on those strengths.

However, to elevate your review even further, you could more explicitly frame your praise in terms of the rubric's language. For instance, instead of simply saying "Definitions are presented with clarity," you might say, "The creator demonstrates a strong command of Conceptualizing by Classifying, as evidenced by their clear and well–aligned definitions of terms such as 'differentiated instruction' and 'equity–focused pedagogy.'" This would make the alignment between your comments and the rubric even more transparent.

Additionally, you might consider briefly acknowledging how the work reflects integration of the Learning by Design framework (e.g., multimodality, recursive feedback), since this rubric specifically includes this as part of its evaluative lens. Doing so would demonstrate your awareness of the overarching pedagogical context and further reinforce your analysis of the work's strengths.

Overall, your review is highly effective in highlighting the creator's successes while using a respectful and constructive tone.

**Rating: 4**

Let me know if you'd like a deeper explanation or examples for strengthening the alignment with rubric language—I'd be glad to help!